\documentclass[conference]{IEEEtran}
\IEEEoverridecommandlockouts
\usepackage{cite}
\usepackage{amsmath,amssymb,amsfonts,bbm,bm}
\usepackage{algorithmic}
\usepackage{graphicx}
\usepackage{textcomp}
\usepackage{xcolor}
\usepackage{pbox}
\usepackage{booktabs}
\usepackage{transparent}
\usepackage{watermark}

\def\BibTeX{{\rm B\kern-.05em{\sc i\kern-.025em b}\kern-.08em
    T\kern-.1667em\lower.7ex\hbox{E}\kern-.125emX}}
\begin{document}

\title{Text-to-Events: Synthetic Event Camera Streams from Conditional Text Input\\
\thanks{This work was partially supported by the Swiss National Science Foundation
 CA-DNNEdge  (208227).}
}

\author{\IEEEauthorblockN{Joachim Ott, Zuowen Wang, Shih-Chii Liu}
\IEEEauthorblockA{\textit{Institute of Neuroinformatics} \\
\textit{University of Z\"urich and ETH Z\"urich}\\
Zurich, Switzerland \\
\{jott, zuowen, shih\}@ini.uzh.ch}
}
\maketitle

\thispageheading{\vspace{-8mm}   \centering\transparent{0.3}This paper has been accepted to the\\ Neuro Inspired Computational Elements Conference (NICE), La Jolla, California, United States of America, 2024.}

\thiswatermark{

\put(0,-730){\parbox{\textwidth}{%
        \transparent{0.5}\textcopyright 2024 IEEE. Personal use of this material is permitted. Permission from IEEE must be obtained for all other uses, in any current or future media, including reprinting/republishing this material for advertising or promotional purposes, creating new collective works, for resale or redistribution to servers or lists, or reuse of any copyrighted component of this work in other works.}
    }%
    }

\begin{abstract}
Event cameras are advantageous for tasks that require vision sensors with low-latency and sparse output responses. However, the development of deep network algorithms using event cameras has been slow because of the lack of large labelled event camera datasets for network training. 
 This paper reports a method for creating new labelled event datasets by using a text-to-X model, 
where X is one or multiple output modalities, in the case of this work, events. 
Our proposed text-to-events model 
produces synthetic event frames directly from text prompts.
It uses an autoencoder which is trained to produce sparse event frames representing event camera outputs. 
By combining the pretrained autoencoder with a diffusion model architecture, the new text-to-events model is able to generate smooth synthetic event streams of moving objects. The autoencoder was first trained on an event camera dataset of diverse scenes. In the combined training with the diffusion model, the DVS gesture dataset was used. We demonstrate that the model can generate realistic event sequences of human gestures prompted by different text statements.
The classification accuracy of the generated sequences, using a classifier trained on the real dataset, ranges between 42\% to 92\%, depending on the gesture group. The results demonstrate the capability of this method in synthesizing event datasets.
\end{abstract}    


\section{Introduction}
\label{sec:intro}

The dynamic vision sensor (DVS), or event camera, was first introduced in~\cite{lichtsteiner2008}. Different from a frame-based camera which operates at fixed frequency synchronously, the DVS asynchronously produces a sequence of events that logs brightness changes at individual pixel locations. 
Event cameras have the advantages of high-dynamic range, high temporal resolution, sparsity in the output data, and low power consumption. These attributes make event cameras suitable for applications such as high-speed robotics~\cite{Delmerico19icra, dvs_robotics, roshambo, deblur}, computational photography~\cite{Mei_2023_CVPR,sun2022event}, and space observation~\cite{star_tracking, nicholas_ralph, brian_space1, brian_space2, brian_space3}; and scenarios with difficult lighting conditions and fast moving objects~\cite{threeET, wang2024ais_event}.
Data collection for event cameras has been a long-standing challenge. Because the camera shows its best advantage under difficult recording conditions, oftentimes it is difficult or even impossible to obtain high-quality ground truth with human labeling.

Researchers rely on event datasets~\cite{amir2017low,xu2016msr,mueggler2017event} to train new algorithms or to improve existing ones for event camera inputs. Such datasets are few in number, below 1\%~\cite{cv:dataset:Fisher:2020,hu2020learning} of well-known computer vision datasets
used for training deep networks. Although some methods can overcome the scarcity and size of event datasets such as the use of unsupervised domain adaptation methods~\cite{hu2020learning,messikommer22domain}, progress in algorithm development will depend on the availability of even more datasets targeted at the desired application.
Collecting a real dataset can be slow and demanding. Although one can use an event simulator~\cite{esim}, it requires the setup of a virtual environment. Converting videos to events is possible~\cite{zhang2023v2ce,hu2021v2e,gehrig2020video}, but this limits the use of these simulators to scenarios where intensity-based cameras 
can record high-quality videos. For fast prototyping, a prompt-based system for event synthesis would benefit the research community for event-based vision. 

So far there is no reported text-to-events system that allows prompt-based generation of synthetic event streams representing motion. To the best of our knowledge, only one other work reported a text-based event synthesis method which produces an event frame from a single image and not
of 
a smooth motion event sequence \cite{zhou2023clip}. Since event cameras detect changes in brightness, their main application domain is in non-static scenes, where the events are generated by either camera movement or object movement within the view of the camera. A useful prompt-based system should 
output good motion representation in the event stream.\\

The contributions of this work are as follows:
\begin{enumerate}
    \item A new method of creating labelled event camera datasets of smooth motion using text prompts as described in Section~\ref{sec:methods}.
    \item A text-to-events model that generates event frames directly without an intermediate step of generating intensity frames.
    \item A novel method for training an autoencoder to produce sparse event frames representative of event camera outputs as detailed in Section~\ref{sec:methods:autoencodr_train}.
\end{enumerate}

We demonstrate our method by training our model to synthesize event streams of human gestures from provided text prompts. These synthesized streams are evaluated both quantitatively and qualitatively in the results.

The remainder of the paper is organized as follows: 
In Section \ref{sec:related_work} we show how our work relates to other developments, followed by a detailed description of our method in Section \ref{sec:methods} and the experimental results in Section \ref{sec:experiments}. Finally, we discuss the limitations in Section \ref{sec:limitations} and present further developments in Section \ref{sec:conclusion}.

\begin{figure}[tbp]
  \centering
   \includegraphics[width=0.8\linewidth]{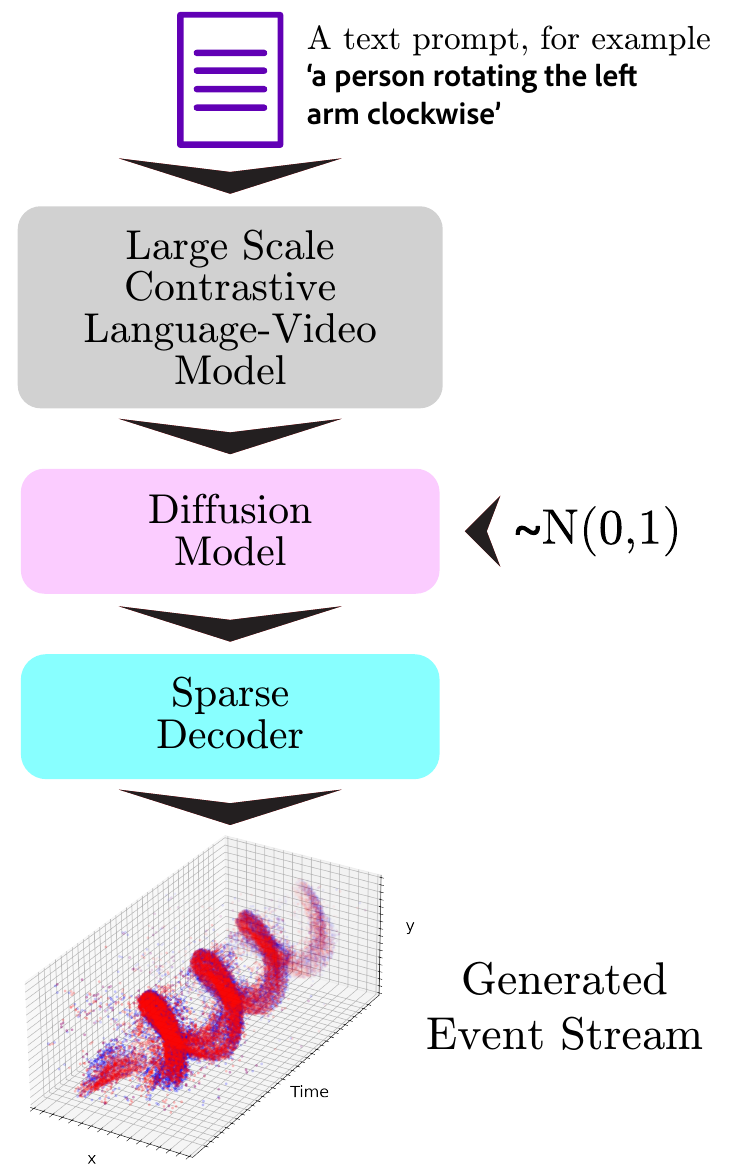}

   \caption{Overview of the entire pipeline. From top to bottom: A text prompt is encoded via a pretrained large scale contrastive language-video model. This embedding is used as conditional input to a diffusion model. The output of this model is fed to a special trained sparse decoder, that outputs event frames representing time-binned event counts. Via Bernoulli sampling from these event frames  ON and OFF event streams of a smooth gesture motion are constructed. }
   \label{fig:full_model_pipeline}
\end{figure}

\section{Related Work}
\label{sec:related_work}
Real event datasets frequently used for developing algorithms include the 
DVS128 gesture dataset \cite{amir2017low}, DAVIS 240C \cite{mueggler2017event}, DDD20~\cite{Hu2020-ddd20} which includes both frames and events, and the Prophesee Gen1~\cite{prophesee_gen1} which only contains events.
While most are on the smaller side by today's standards, for example, DVS128 gesture has 1342 clips of around 6 seconds, there are a few large ones such as the Gen1 dataset with a total length of 39 hours. The samples in these datasets are very sparse, for example the `boxes 6dof' sample of DAVIS 240C is 59.8 seconds long and contains only approx. 133 million
events. 
Compared to the events from a dense full frame representation of approx. 59 million 
$\mu$s  
(the timestamp resolution of the events) multiplied by $240 \times 180$ (the pixel resolution of the DAVIS camera) leading to a total of $2.5\times 10^{12}$ values, this is only a fill rate of 0.005\%.

\subsection{Synthetic Event Generation}
 Video-to-event toolboxes such as rpg$\_$vid2e~\cite{gehrig2020video} and v2e~\cite{hu2021v2e} synthesize DVS events from videos via different modeling details and generation methods. 
The rpg$\_$vid2e toolbox assumes an idealistic model of DVS pixels, that is, the
pixel bandwidth is always large enough for the input video rate.
Since synthetic slow motion requires sharp,
high-quality intensity frames with almost no motion blur, the rpg$\_$vid2e toolbox allows the simulation
of idealized DVS pixels under good lighting, which does not correspond to realistic real DVS pixel events under difficult illumination conditions. v2e allows the generation of simulated datasets covering a range of lighting conditions with the help of a more realistic model of the DVS pixel taking into account temporal noise, leak events, and pixel threshold mismatch.
Both toolboxes usually rely on pretrained frame-interpolation models for higher temporal resolution of the source video, before converting this new sequence of frames to intensity maps and finally, an event stream using the DVS pixel models. These conversion toolboxes are convenient to use but they all require real high-quality videos first. Another option is event camera simulators such as ESIM~\cite{esim} that employ virtual event cameras in a computer graphics based virtual environment. These simulators require a heavy virtual setup to be built and properly configured before the event sequences can be generated.

The only text-to-events model, EventBind~\cite{zhou2023clip}, can be used to generate events from text but is limited to mostly static scenes without motion.

\subsection{Text-to-(Modality)}
Recently, new large language models, often in combination with other model components such as diffusion models~\cite{sohl2015deep,ho2020denoising,yang2022diffusion,croitoru2023diffusion} have led to a surge of reported methods that allow synthetic generation of increasing quality samples.
The output of these models is often referred to as `AI generated X', where X can be any modality, for example, music~\cite{copet2023simple}, human motion~\cite{karunratanakul2023guided,azadi2023make}, images~\cite{ramesh2022hierarchical,rombach2022high}, and videos~\cite{luo2023videofusion,wang2023lavie,ho2022imagen}. These models typically leverage pretrained models providing embeddings, for example from a joint video and text embedding space~\cite{xu2021vlm,ni2022expanding,wang2023all}, to enable text-to-video pipelines.

These pipelines usually create low frames-per-second~\cite{luo2023videofusion} clips, and often struggle with preserving object appearance over time~\cite{chai2023stablevideo}. This constant change of object appearance, especially prominent as noisy object surfaces and outlines, make these synthetic video clips unsuitable to be converted to events via a video-to-events toolbox.

\section{Methodology}
\label{sec:methods}
In this section we describe the different components of our model, and the methods used for data preprocessing, model training and benchmarking.

\subsection{Pipeline Overview}\label{sec:methods:overview}
As can be seen in Fig.~\ref{fig:full_model_pipeline}, our text-to-events pipeline consists of 3 components: 1) a pretrained large scale contrastive language-video model (VLM), 2) a sparse autoencoder, and 3) a diffusion U-Net. Common event datasets \cite{amir2017low,mueggler2017event} contain short sequences, but because of the high temporal resolution, the timestamps of these sparse inputs often span over millions of $\mu$s.
We therefore reduce the long temporal dimension of the input to a lower-dimensional yet informative representation 
by first filtering the input (see Section~\ref{sec:eventfilter}) and then creating event frames (see Section~\ref{sec:eventrep}). Finally, as is common in latent diffusion model training~\cite{yang2022diffusion,croitoru2023diffusion}, we train an autoencoder to provide further dimensionality reduction and the necessary latent space for the diffusion model. 

\subsection{Event Filtering}
\label{sec:eventfilter}
The DVS event stream consists of a sequence of
events. Each
event, marked by index $i$ in an event stream, is expressed as
$e_i = (x_i, y_i, t_i, p_i)$, where $(x_i, y_i)$ signifies the pixel location,
$t_i$ is the timestamp, and $p_i \in \{\pm 1 \}$ is the polarity or
direction of the brightness change.

\begin{figure}[htbp]
  \centering
   \includegraphics[width=0.9\linewidth]{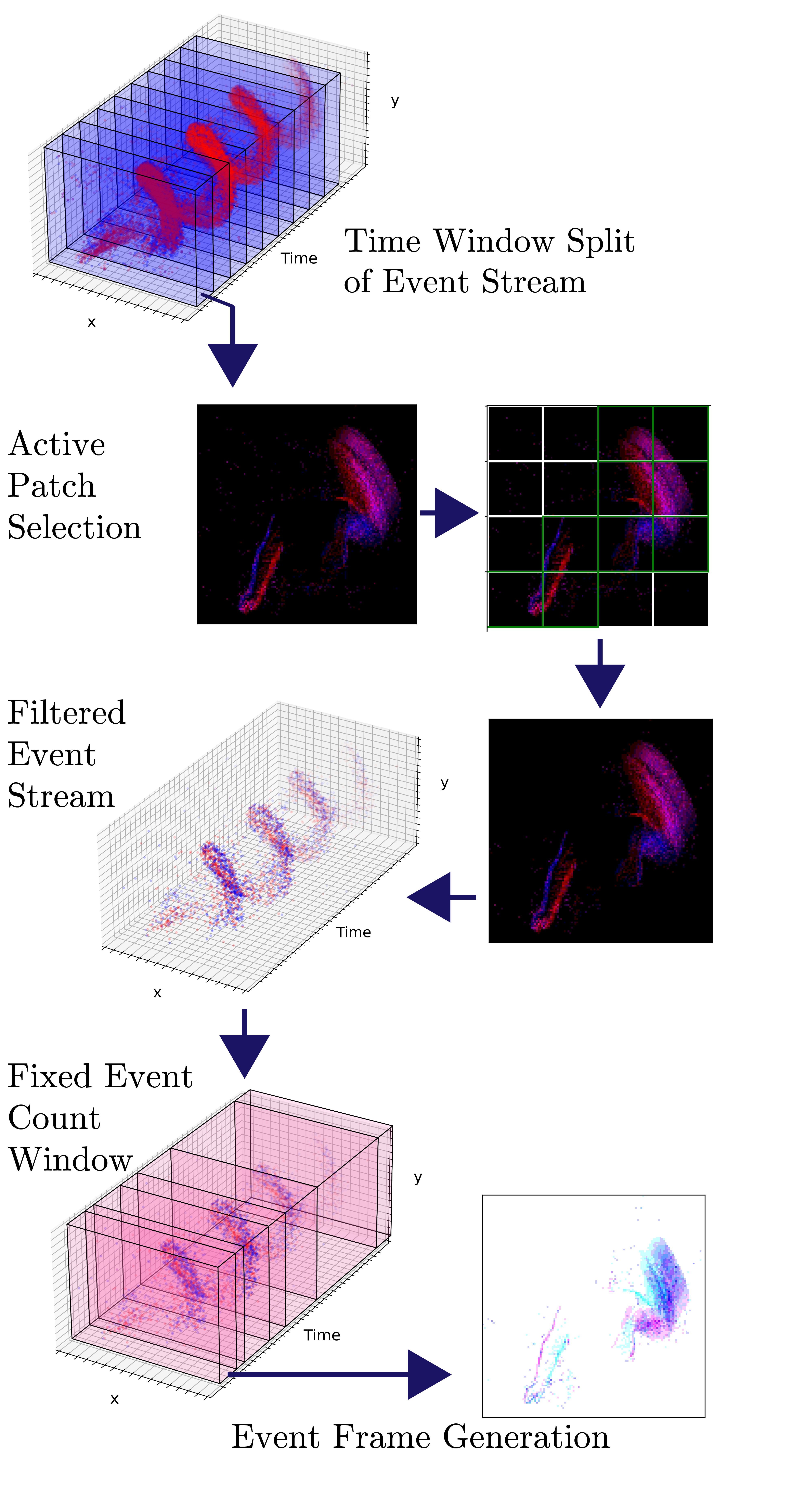}

   \caption{Data Preprocessing: Event streams from the datasets are split into smaller slices of a fixed time window length. Within these slices, a patch grid in the image dimension is overlaid. If the count of all ON and OFF events within a given patch is below a threshold these events get removed from the event stream. The filtered event stream is then split into slices of a fixed event count and each slice is converted to an time binned event frame in the format of a spatio-temporal voxel grid.}
   \label{fig:event_preprocessing}
\end{figure}

Real-world event streams frequently contain pixel noise \cite{hu2021v2e}. Various advanced and often computationally demanding methods for event de-noising and super resolution (increasing temporal and/or spatial dimension) were proposed; an overview can be found in \cite{duan2021eventzoom}. A simpler method employed by other works is called time-slice active patch selection \cite{wang2022exploiting,sabater2022event}. As shown in Fig.~\ref{fig:event_preprocessing}, if a patch within the pixel grid, either of the raw events or a form of event-frame representation, has less than a threshold value of events within a time window, the events are removed during preprocessing. As shown in the lower part of the figure, we split the filtered event stream into fixed event count slices and convert each slice to the event representation described next. 

\subsection{Event Representation}

\label{sec:eventrep}
To train an ANN model, the format of events $e_i = (x_i, y_i, t_i, p_i)$ is usually converted into some form of sparse or dense tensor. A recent overview of event representation methods was provided by \cite{zubic2023chaos}. In our work, we choose the separate polarity time-binning method for a spatio-temporal voxel grid, where 
positive and negative events are handled separately. Each type of event at a specific x and y location is time-binned into $C$ channels, where each bin is of a specific time interval. The resulting representation is therefore a spatio-temporal voxel grid. The events in each time bin are summed up. Note that since we accumulate the number of events thus for the negative polarity channels, the values in this grid are also all positive. In addition, we use a fixed event count window to extract events from the filtered event stream 
and then convert to the voxel grid format. The timestamps of the events in each extracted event count window are normalized such that the first event in each window occurs at $t=0$ and the last at $t=1$, and the time interval of each bin corresponds to $1/C$.

\subsection{Iterative Autoencoder Pretraining}\label{sec:methods:autoencodr_train}
\begin{figure}[tp]
  \centering
   \includegraphics[width=1.\linewidth]{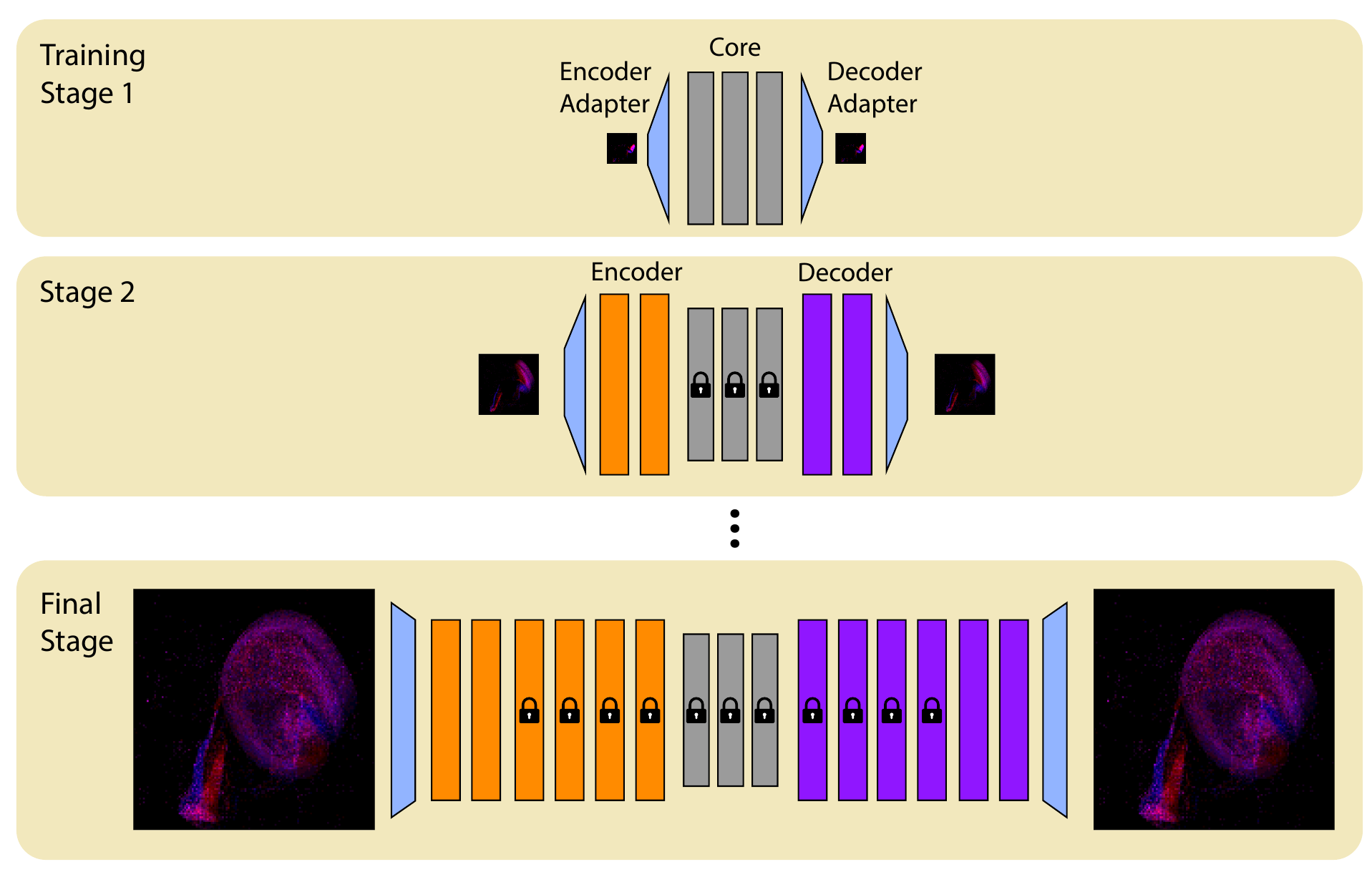}

   \caption{Iterative autoencoder (AE) training: Training starts at stage 1 with a low resolution of 8x8 pixels and the AE core is trained together with a adapter input and adapter output layer. These adapter layers map the input resolution to the input size of the next layer and last layer output to the output resolution (same as input resolution). In the next stage, the core layer parameters are frozen, the adapter layers get discarded, and new encoder and decoder layers are added, together with a new pair of adapter layers. The input event frames are $2\times$ the resolution of the previous training stage. This iterative process continues until the final $128\times128$ resolution is reached. An additional fine-tuning step with training on all parameters may be done as well.}
   \label{fig:aemodel}
\end{figure}

For our pipeline to
support both the encoding and decoding for the event-frame format,
we need a generative component, hence we chose an autoencoder. Although existing works may offer (pretrained) encoders, they are usually not trained on an event decoding task. For example, the E2Vid~\cite{rebecq2019high} encoder does not preserve motion representation since it is optimized for semantic information preservation to enable frame reconstruction \cite{messikommer2023data}. The same can be observed from the autoencoder of \cite{vemprala2021representation}, because their model employs max pooling over the time axis. Another requirement placed on our pipeline is the use of a lightweight autoencoder because diffusion model training requires considerable resources. We therefore build an autoencoder using a combination of CNN and Dense layers. Training such a model on sparse input (and output) can lead to training collapse, for example, the posterior collapse in a VAE \cite{he2019lagging}.
We chose to develop a training technique
similar to progressive VAE training \cite{utyamishev2023multiterminal}, which is one of the methods used to overcome this
collapse~\cite{he2019lagging,bredell2023explicitly,utyamishev2023multiterminal}.
As shown in Fig.~\ref{fig:aemodel}, our autoencoder consists of multiple layers for both the encoder and decoder. They are connected via a core consisting of dense neural network layers. 

\textbf{Iterative training}
This training consists of different stages where both the image resolution and the number of encoder and decoder layers are increased per stage. As shown in Fig.~\ref{fig:aemodel}, in the first stage, the core layers get an added input and output adapter layer. Core layers are multiple fully connected layers, and the adapter layers are convolutional layers. After training in this configuration, the input and output adapter layers are discarded and the core layer parameters frozen. In the next stage, the image resolution is doubled per image dimension (hence 4$\times$ pixels in total), and both an encoder and decoder block of convolutional layers are added, along with a new set of input and output adapter layers. The training stages further increase in this fashion until the final image resolution is reached. As a final step, one may perform a fine-tuning on the network with the full image resolution and all model parameters unfrozen, hence trainable.

\subsection{Warm-Up for Full Event Occupancy}
\begin{figure}[tbp]
  \centering
   \includegraphics[width=1.\linewidth]{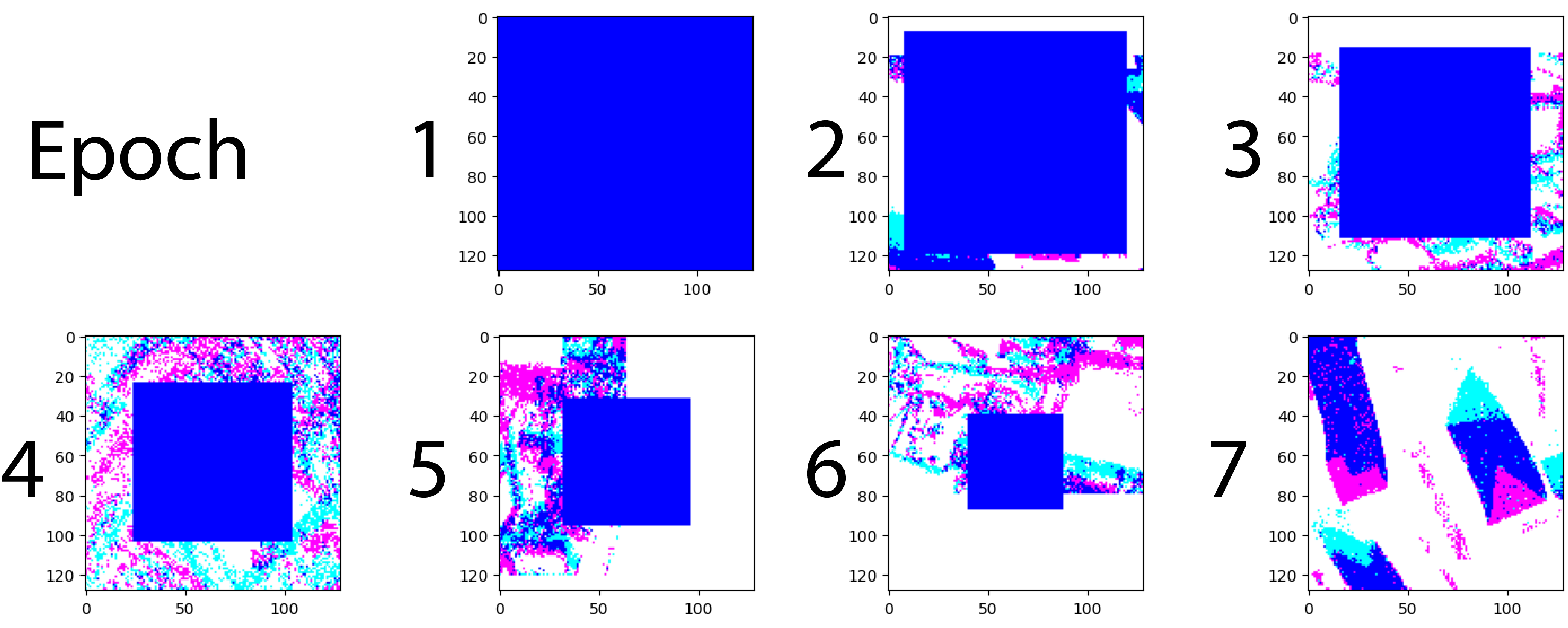}

   \caption{Warm-up for full event occupancy during autoencoder training: To reduce the probability of a collapse of the decoder to only output zero events everywhere, we set a decreasing number of values in a training batch to the maximum value in said batch. In the first epoch, here depicted for a $128 \times 128$ image resolution, all values are set to the maximum. In epoch 2 an outer rim of each event frame in the form of a voxel grid stays the original values. The number of values set to the batch maximum value is further decreased in the following epochs. Starting from epoch 7, no voxels are set to the maximum value.}
   \label{fig:occupancy}
\end{figure}

So that the decoder does not
collapse and produces an all-zero output, we employ a warm-up data augmentation regime as depicted in Fig.~\ref{fig:occupancy}. We augment the spatio-temporal voxel grids in the following way:
For the first $q$ training epochs, we set the values of each training sample in a given training batch to the maximum value of the batch. In the first epoch all voxel grid values are replaced with the maximum batch value, in the next epoch, all values are set to the maximum except for an outer rectangular area where the original values of each grid are maintained, and in each subsequent epoch, the area set to the maximum batch value shrinks towards the center as shown in Fig.~\ref{fig:occupancy}. In detail:
any pixel $X_{i,j}$ at coordinate $(i,j)$ in an voxel grid $X$, has the value, $x_{i,j}$, during warm-up at $n_{\text{th}}$ epoch if:
\begin{gather}
z= \max\left(\left\lfloor \frac{{8\cdot M_s}}{{M}}\right\rfloor, 1\right) \cdot (n - 1)\\
    x_{i,j} =
    \begin{cases} 
\text{max}(X), & \forall (i,j) \in [z,M_s-z) \times [z,M_s-z) \\
x_{i,j}, & \text{otherwise} 
\end{cases}
\end{gather}
where $z$ is the length of the fill-in border, $M_s$ is the image dimension of the current training stage and $M_s \leq M$; and $\times$ represents the Cartesian product. $[z,M_s-z) \times [z,M_s-z)$ gives the data augmentation region (marked with blue in Fig.~\ref{fig:occupancy}) which will be assigned with the maximum value of the input. This warm-up prevents the model from collapsing due to the sparsity of event frames.

\subsection{Event Population Regularized Loss}
The time-binned representation of an event sequence from a sample is described by an event frame in the format of a spatio-temporal voxel grid $X \in \mathbb{R}^{2C \times {W} \times {H}}$, where $C$ is the number of voxel channels containing the per time window bin count for the positive and negative events respectively.
 $W$ and $H$ define the voxel grid 
width and height.

\textbf{Mean squared error}
The base loss function for the training pipeline is the channel-wise mean squared error
$\mathcal{L}_{\text{MSE}} \in \mathbb{R}^{2C} $, for given a sample $X$ and its reconstruction $\hat{X}$ it is defined as:
\begin{gather}\mathcal{L}_{\text{MSE}}=\text{MSE}(X,\hat{X})={\frac{1}{W \cdot H}}\sum_{i=1}^{W}\sum_{j=1}^{H}\left(X_{i,j}-{\hat{X}_{i,j}}\right)^{2}
\end{gather}

\textbf{Sparse population regularization} To encourage sparsity in the generated event stream while also avoiding decoder collapse, the base loss obtained using the reconstruction gets scaled. The amount of scaling of $\mathcal{L}_{\text{MSE}}$ depends on how the output frame sparsity is different from the given input event frame sparsity. Less reconstructed events compared to the input increases the loss more strongly, since this is closer to all-zeros output of a decoder collapse.
Analogous to the regularization method introduced in \cite{utyamishev2023multiterminal}, we use a Heaviside step function $\mathbbm{1}(\cdot)$~\cite{abramowitz1972handbook,weisstein2002heaviside} to first determine the number of non-zero voxels in the input voxel grid as well as in the reconstruction grid. Then we measure their sparsity distance by subtracting these two counts of non-zero voxel values. The larger the distance, the more dissimilar they are in terms of events statistics. The voxel channel-wise distance is defined as:
\begin{gather}\text{distance}= \sum_{i=1}^{W}\sum_{j=1}^{H}\mathbbm{1}(\hat{X}_{i,j})-\sum_{i=1}^{W}\sum_{j=1}^{H}\mathbbm{1}(X_{i,j}). 
\end{gather}. 

Along with the proposed values in~\cite{utyamishev2023multiterminal} for the hyperparameters, $k_{\text{err}}=1e^2$ and $k_{\text{subopt}}=1e^{-3}$, we define the step of the loss as:
\begin{gather}
\text{step}=k_{\text{err}}\cdot \text{sign}(\text{distance}-1)+1.
\end{gather}
The complete channel-wise sparse population regularization scaling term is
\begin{gather}
\mathcal{S}_{\text{sparse}}=1+ k_{\text{subopt}} \cdot \text{step} \cdot \text{distance}.
\end{gather}
In our application of this loss, $k_{\text{err}}$ can be interpreted as the penalty factor for too few events and $k_{\text{subopt}}$ for too many events. 

\textbf{Channel weight}
Depending on a particular event voxel grid and even over a dataset, 
there can be an imbalance of the number of ON and OFF events. Additionally, if multiple time bins in the form of voxel channels are used per polarity, there can also be an imbalance in the number of events (hence non-zero values) between different temporal bins. If a dataset has a bias towards, for example ON events, this can cause issues in training an autoencoder. To reduce the effect of this bias, we apply a weighted loss function as follows:
First, we determine the reverse proportion $C_{\text{rp}} \in \mathbb{R}^{2C}$ of each voxel channel:
\begin{gather}
        C_{\text{rp}; c}= 
        1-\frac{\sum_{i=1}^{W}\sum_{j=1}^{H}X_{c,i,j}}{\sum_{c=1}^{2C}\sum_{i=1}^{W}\sum_{j=1}^{H}X_{c,i,j}}
\end{gather}
The channel weight  $\bm{C_{\text{w}}} \in \mathbb{R}^{2C}$ with a weight factor $C_{\text{w}; c}$ for each channel is defined as:
\begin{gather}
        C_{\text{w}; c}= \text{max}(c_{\text{min}}, C_{\text{rp}; c} )
\end{gather}
with $c_{\text{min}}$ set to 0.1 in all experiments.

\textbf{Total loss}
By combining all components, we get
\begin{equation}
    \mathcal{L} = \sum_{2C} \mathcal{L}_{\text{MSE}} \cdot \mathcal{S}_{\text{sparse}} \cdot \bm{C_{\text{w}}}
\end{equation}

\textbf{F1 for evaluation} We evaluate the autoencoder using the F1 scores:

\begin{equation}
    F_1 =  \frac{2 \text{TP}}{2\text{TP} + \text{FP} + \text{FN}}
\end{equation}
where TP, FP, FN stands for true positive, false positive, and false negative respectively, and calculated on the non-zero pixels of ground truth and the reconstruction result.

\subsection{Diffusion Model with Conditional Text Input}
We train a latent diffusion model~\cite{rombach2022high} (DM) to generate latent representations of event sequences from VLM encoded text prompts. These latent representations are subsequently decoded by the decoder part of the autoencoder. 
From the decoder output, Bernoulli sampling is used to generate event sequences.
For training, to convert an event sequence to DM input, we encode all fixed event count slices of an event sequence using the autoencoder, and stack these encodings. The input to the DM, which we call $x_0$, is therefore a 2D matrix.\\

\textbf{Working principles: }Diffusion models assume some form of forward function that adds noise to input \cite{peebles2023scalable}. The DM then learns the iterative reverse process of de-noising. Besides the noisy input, these models may also get a conditional input, e.g. an embedding of text or a class label. The noise is assumed to be Gaussian, but many different variations of DMs and training techniques exists~\cite{yang2022diffusion,croitoru2023diffusion}. 
For our purposes, we can define the objective with conditional input more formally~\cite{saharia2022photorealistic} as 
\begin{equation}
    \mathbb{E}_{\bm{x},\bm{c},\bm{\epsilon},t}\left[w_t\|  f_\theta(\gamma_t\,\bm{x}+\sigma_t\bm\,{\epsilon},\bm{c})- \bm{x}\|_{2}^{2}\right]
\end{equation}
where $(\bm{x},\bm{c})$ is a pair of input data and condition, $\bm{\epsilon}\sim\mathcal{N}(0,\mathbf{I})$, $t\sim\mathcal{U}([0,1])$ denotes a step in the noise process, $\gamma_t$ a noise level indicator, and $\sigma_t$ and $w_t$ are additional functions of $t$ that influence de-noising quality. 
$f_\theta$ learns to denoise $\bm{z}:=\gamma_t\,\bm{x}+\sigma_t\,\bm{\epsilon}$, hence steps in the denoising process are functions of the data predictions $\hat{\bm{x}}:=f_\theta(\bm{z}_t,\bm{c})$.
By using classifier-free guidance~\cite{ho2022classifier} and training on conditional and unconditional inputs simultaneously, we can use the simplified technique from~\cite{nichol2021improved} such that the model becomes a noise predictor, MSE is used as the loss function, and we adjust the prediction of $\bm{x}$ according to~\cite{saharia2022photorealistic}: 
\begin{equation}
    (\bm{z}_t-\sigma\Tilde{\bm{\epsilon_\theta}})/\gamma_t
\end{equation}
where
\begin{equation}
    \Tilde{\bm{\epsilon_\theta}}(\bm{z}_t,\bm{c})=w\bm{\epsilon}_\theta(\bm{z}_t,\bm{c})+(1-w)\bm{\epsilon_\theta(\bm{z}_t)}
\end{equation}
The term $\bm{\epsilon}_\theta:=(\bm{z}_t-\gamma_tf_\theta)/\sigma_t$, $\bm{\epsilon_\theta(\bm{z}_t)}$ is the unconditional input, and $\bm{\epsilon_\theta(\bm{z}_t,\bm{c})}$ is the conditional input. $w$ is called the \textit{guidance scale} or \textit{guidance weight} and adjusts the impact of the conditional input. We set $w=7.5$ in all our experiments.
\section{Experiments}
\label{sec:experiments}

In this section we explain the details of our experiments, as well as their results and interpretation.

\begin{figure}[tbp]
  \centering
   \includegraphics[width=1.\linewidth]{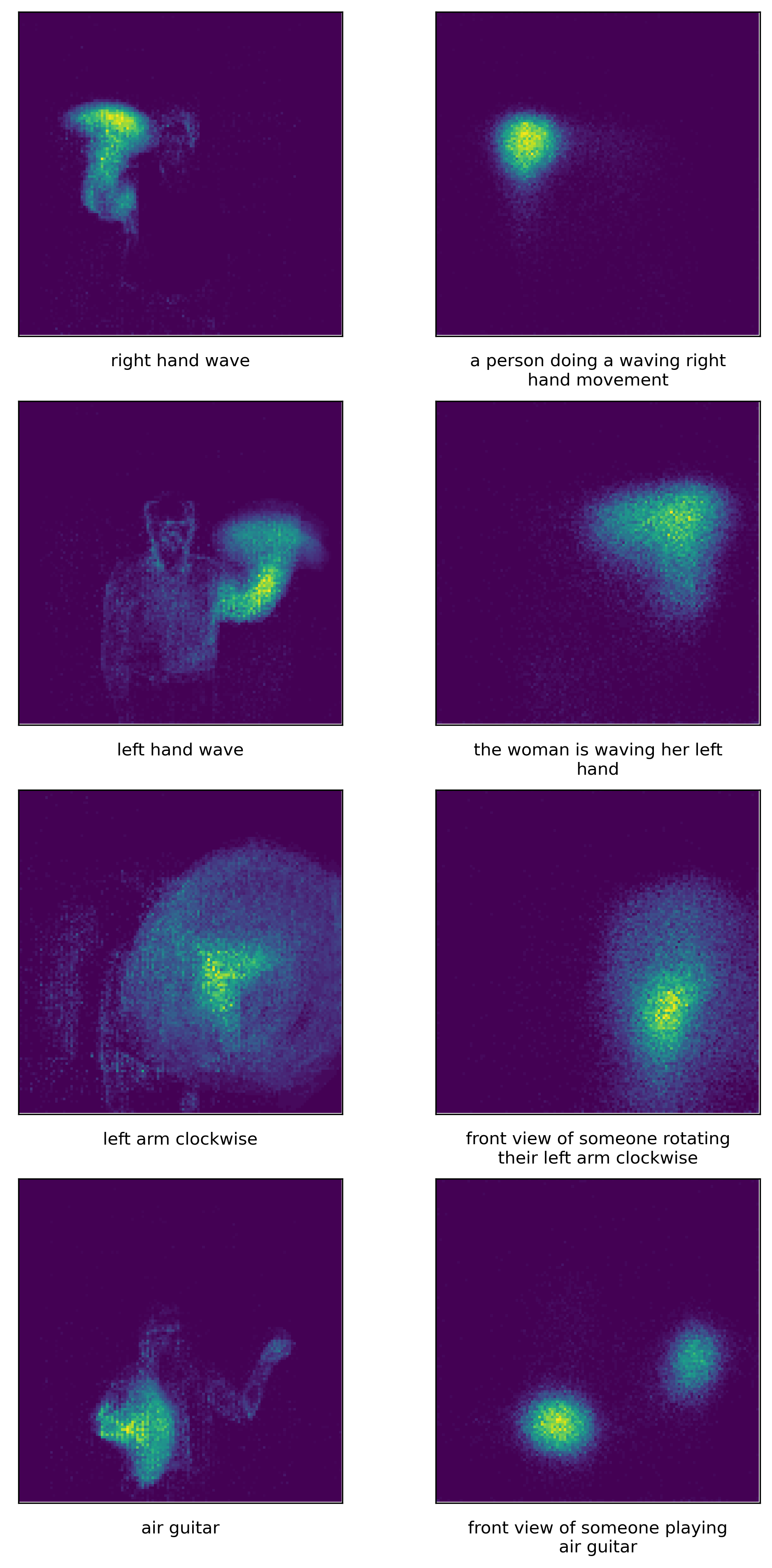}

   \caption{Sum of ON events over the first 100 spatio-temporal voxel grids. Ground-truth (left column) dataset samples, each with its class label, and generated event streams (right column) and the corresponding prompt used for generation. Our model correctly emphasizes event generation in the relevant motion trajectory areas. }
   \label{fig:diff_gen_subset}
\end{figure}

\begin{figure}[tbp]
  \centering
   \includegraphics[width=1.\linewidth]{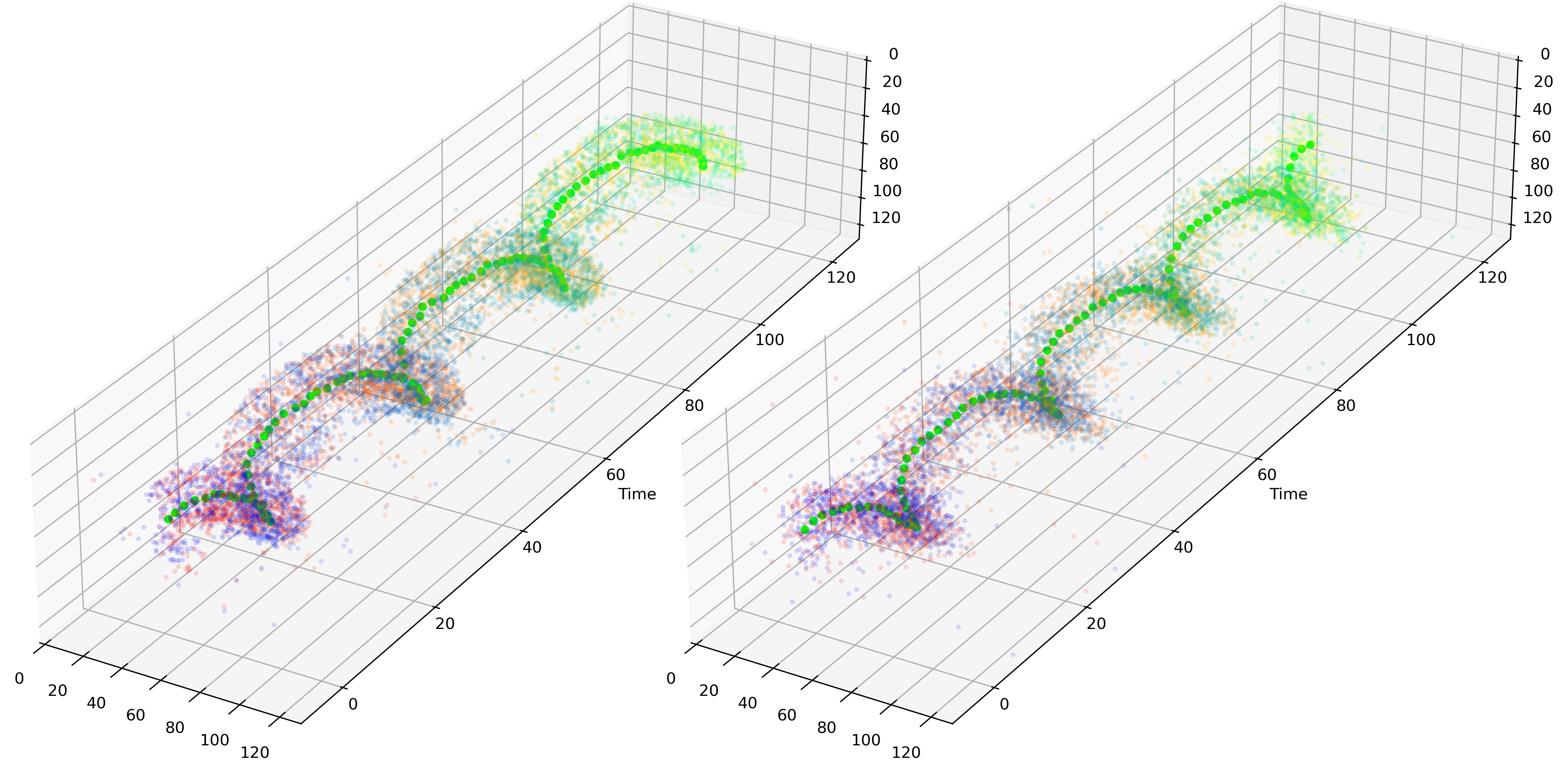}

   \caption{2D space-time visualization of a real sample (left) of 'right arm counter-clockwise' from the DVS128 gesture dataset, and a generated sample (right) using the prompt 'a man appears to be rotating his right arm counter clockwise'. ON events are in red-to-yellow gradient over time, OFF events are in blue-to-dark-green over time. For better motion trajectory visualization, the mean event position per step is highlighted in light green. Our model is able to generate a smooth trajectory of ON and OFF events. }
   \label{fig:diff_spacetime}
\end{figure}

\textbf{Datasets} We use the DVS128 gesture dataset~\cite{amir2017low}, for the diffusion and classifier training; and the DAVIS 240C dataset~\cite{mueggler2017event} for pretraining of the autoencoder. We do not use the `background' and `random' class (class labels 0 and 11), hence we have 10 remaining classes. We hand-crafted 36 different prompts for each of of the 10 DVS128 classes. For example, besides the provided label `clapping hands' for the hand clapping class we devised among many other variations `a person is clapping hands' and `front view of someone clapping hands'. We then generated the embeddings of each prompt using the VLM and further created a UMAP~\cite{mcinnes2018umap} clustering of these embeddings. This clustering process identified prompt variations with words, such as `toon' that are not within the training vocabulary of the VLM and caused all prompts containing this word to collapse into a similar embedding space area. These prompts are then filtered out.
Dataset splits: DAVIS240C does not provide a split, and we use it for training only, hence we use the entire dataset. DVS128 provides a train and test split. 

\textbf{Autoencoder}
The core of the autoencoder consists of a flattening layer followed by 3 fully connected layers with GELU~\cite{hendrycks2023gaussian} activation and Dropout~\cite{srivastava2014dropout} in between layers. In the final stage, the encoder consists of 6 blocks each with a 2D GELU convolution~\cite{lecun1989backpropagation} layer and 2D maximum pooling layer. The last block has no maximum pooling layer. The decoder consists of an unflattening layer followed by 5 blocks of 2D GELU convolution and upsampling layers. The final layer is 2D convolution with sigmoid activation.

\textbf{Diffusion conditional U-Net} We use the 2D conditional UNet with 6 input and 6 output blocks each. The 6 input blocks consist of 4 ResNet downsampling blocks, one with spatial self-attention, and finally another downsampling block. The block sizes are $128, 128, 256, 256, 512, 512$. The other side of the U-Net has the same layer configuration in reverse. Cross-attentional input from the VLM has a dimension of $77 \times 768$. In practice, for every batch sample, we also feed an empty text embedding of the same size as the unconditional input.
The encoded event sequences, which are concatenated outputs from the encoder part of the autoencoder have just 1 channel, hence the input to the diffusion model has the shape $256 \times 256 \times 1$.

\subsection{Pretraining on DAVIS 240C Dataset}
\label{subsec:davis240c_pretraining}
For the active patch filtering described in Section~\ref{sec:eventfilter}, we use a window length of 20 milliseconds, patches of 8$\times$8 pixels, and an event count threshold of 7 events (if there are $<$ 7 events in a patch, we discard all events in this patch of this time window). We convert the filtered event streams to batches of event frames and train the autoencoder using the technique in Section~\ref{sec:methods:autoencodr_train}.

For evaluation of the reconstruction performance of the autoencoder, we shuffle the test split of the DVS128 dataset using a fixed random seed and take the first 200 samples as the validation set for evaluation. The main evaluation is done through the F1 score of non-zero elements.

We train the autoencoder in different configurations to find a suitable version for a representation-rich latent space.
As shown in Table~\ref{table:ae_pretraining_davis240c_results}, the best performing autoencoder, EAE-1F, uses only 1 bin per polarity. Channels per polarity is corresponds to $C$ as described in Section~\ref{sec:eventrep}, F1 is F1 score of non-zero elements, NI indicate models from runs directly trained on the last stage without iterative training, and Loss is reconstruction loss on the validation set. $C>1$ would result in a higher temporal resolution, however, only runs with $C=1$ (EAE-1, EAE-1F, EAE-1NI) produce qualitatively useful decodings. Adding a finetuning step ('F' at the end of the model name) after the final stage further decreases the loss and improves the F1 score (EAE-1 vs EAE-1F).

During autoencoder training, as described in Section~\ref{sec:methods:autoencodr_train}, we increase the maximum number of events per event frame for each stage. As default we increase from 8$\times$8 pixels
to the final 128$\times$128 pixels
stage in the order 128, 512, 2048, 8192, and finally 16384 events. For the ablation study shown in Table~\ref{ae_pretraining_davis240c_results_ablation}, we also trained with a fixed number of events (128) over all stages (EAE-1-ABL-128), and added a fine-tuning phase (EAE-1-ABL-128F). All experiments were done with 1 channel each for ON and 1 for OFF events. Without increasing the event count when increasing the spatial dimension, the autoencoder fails to learn a working encoding and decoding scheme.


\begin{table}[tbp]
\caption{Impact of Autoencoder Pretraining.  Bold: best F1 score, NI: no iterative training (only last stage), F is finetuned. Details in text.}
\label{table:ae_pretraining_davis240c_results}
\begin{center}
\scalebox{1.0}{
\begin{tabular}{llll}
\multicolumn{1}{l}{\bf \pbox[l]{\textwidth}{Autoencoder \\ Model}} &\multicolumn{1}{c}{\bf \pbox{\textwidth}{Channels per\\ Polarity} } & \multicolumn{1}{l}{\bf F1} & \multicolumn{1}{l}{\bf Loss} \\ 
\midrule
EAE-4F&     4 & 2.353e-05 & 3.799e-05 \\
EAE-2NI &     2 & 2.749e-03 & 1.001e-04 \\
EAE-2&     2 & 2.936e-02 & 1.002e-04  \\
EAE-4 &     4 & 6.147e-02 & 3.8e-05  \\
EAE-1NI &     1 & \textbf{2.044e-01} & 4.21e-02  \\
EAE-1     &1 & \textbf{2.044e-01} & 5.706e-02  \\
EAE-1F     &1 & \textbf{2.044e-01} & 4.126e-02  \\
\end{tabular}
}
\end{center}
\vspace{-6mm}
\end{table}

\begin{table}[tbp]
\caption{Ablation Study. Details in text.}
\label{ae_pretraining_davis240c_results_ablation}
\begin{center}
\scalebox{1.0}{
\begin{tabular}{llllll}
\multicolumn{1}{l}{\bf \pbox[l]{\textwidth}{Autoencoder \\ Model}}  & \multicolumn{1}{l}{\bf F1} & \multicolumn{1}{l}{\bf Loss}  & \multicolumn{1}{l}{\bf Max Events} \\
\midrule
EAE-1F  & \textbf{2.044e-01} & 4.126e-02 &16384\\
EAE-1-ABL-128&6.027e-03 & 3.023e-03                 &  128         \\ 
EAE-1-ABL-128F &6.806e-03 & 3.023e-03      &  128         \\ 
\end{tabular}
}
\end{center}
\end{table}

\subsection{Training}

\subsubsection{Setup and Implementation}
We use the PyTorch~\cite{paszke2019pytorch} framework for all our models. Diffusion models additionally use the Diffusers library \cite{von-platen-etal-2022-diffusers} and configurations from audio-diffusion library~\cite{audio-diffusion}. Some of the event frame processing are heavily modified versions of the original functions released by the authors of \cite{vemprala2021representation}. The VLM is the pretrained language-video model `xclip-large-patch14-16-frames' provided by Microsoft~\cite{XCLIP} and hosted on HuggingFace.

The final stage of the autoencoder has in total 3.3 million parameters, the latent diffusion model has 120.8 million parameters, and the pretrained VLM has 12.3 million parameters.

The autoencoder was trained for 2000 epochs per stage, and another round of 6100 epochs of fine-tuning all parameters on the 128x128 input, which took about a week. The diffusion model was trained for 2000 epochs, which also took about a week.
For the autoencoder training, we use a single T4 GPU per experiment. For the diffusion pipeline training, we use a single V100 GPU.
All experiments use the AdamW optimizer~\cite{Kingma2014,loshchilov2017decoupled}, a learning rate of $10^{-4}$ for autoencoder and classifier training, and $10^{-5}$ in the diffusion model training.

\subsubsection{Diffusion Training}
 We filter the DVS128 gesture dataset using the same parameters presented in Section~\ref{subsec:davis240c_pretraining}. 
Qualitatively, the generated samples show ON and OFF events of motion trajectories similar to the gesture dataset. Videos generated from the generated event streams show a diversity of speed and location of the gesture movement. Fig.~\ref{fig:diff_gen_subset} shows visualizations of the ON events from the dataset samples and the generated samples. Fig.~\ref{fig:diff_spacetime} shows how the mean pixel location shifts over the course of time in a similar manner for both the real and generated samples.

\subsection{Classification of Generated Samples}

\begin{figure}[tbp]
  \centering
   \includegraphics[width=1.\linewidth]{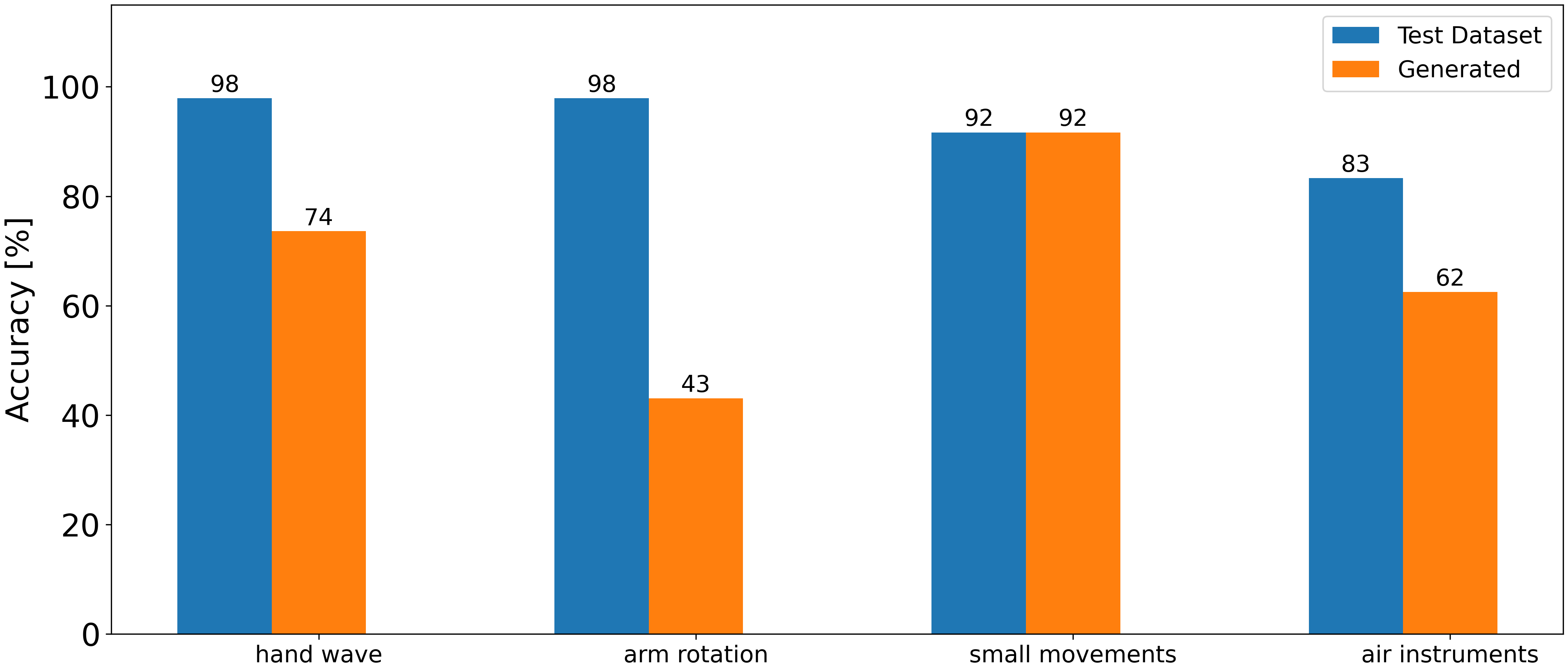}

   \caption{The blue bars are mean classification accuracy values of gesture groups on the DVS128 test set. The overall mean is 93.8\%. In orange are mean accuracy values using the same classifier over all prompt versions of generated samples in each gesture group. }
   \label{fig:diff_accuracy}
\end{figure}

To evaluate the generated samples, we train a classifier on the DVS128 training set. The architecture is the same as the encoder part of the autoencoder. Sequence chunks of fixed event count are encoded. These encodings are stacked to add a temporal dimension. The max and mean along this time axis for each feature dimension is calculated. The max and mean vectors are concatenated and serve as input to the first of 3 subsequent fully connected layers, where the last layer has 10 output units for classification. The classifier is trained with cross entropy loss on the DVS128 training set. Note: these training runs are done from scratch, we do not share any trained weights from the autoencoder training. The classification accuracy on the DVS128 test set is at 93.8\%. The results for different gesture groups are shown in Fig.~\ref{fig:diff_accuracy}. 

To evaluate the generative performance, we generate samples for all 10 classes from all corresponding 36 prompts, hence we have 360 total samples. We then evaluate the class accuracy using the same classifier for the real test set. 
Using the default probability output by the autoencoder to sample events via Bernoulli sampling results in a mean classification accuracy of 57.2\%. If we boost the probability (multiply by at least 3 and clip at 1) of all generated events to 1, we get 62.8\% accuracy. 
This shows that the classifier needs a certain amount of events to predict a class, although random removal of events was part of the data augmentation process during classifier training. The error happens mainly with large scale rotational arm movements which have an average accuracy of only 43\%. 
For the other 3 groups, we get a mean accuracy of 75.9\%. On the real test set, we get 97.9\% for the rotational arm movements and 90.9\% on the other groups. 

A 2D space-time visualization of both a real and a generated sample of such an arm rotation can be seen in Fig.~\ref{fig:diff_spacetime}. If we qualitatively inspect the generated arm rotation samples we can see proper movement, both in terms of which arm and direction of rotation. A new classifier trained on both generated samples and real samples may improve the accuracy scores. 

\subsection{Known Limitations}
\label{sec:limitations}

A major limitation is the diffusion training on gesture descriptions and inputs, hence only allowing for prompts of this type during inference. Though computationally demanding, training on a large video-language dataset with diverse scenes would result in a general-purpose text-to-events model. This would also require conversion of the dataset to events first. Another limitation is the resolution and specificity of the embedding space from the pretrained language model. Future language-video models may provide finer-grained embedding space that is suitable for more prompt specificity. 
Combining our model with a future text-to-video model may enable joint intensity frames and DVS event output as is available from some event camera sensors like DAVIS~\cite{Brandli2014-cs}.
\section{Conclusion}
\label{sec:conclusion}
This paper presented the first text-to-events model capable of generating event streams of smooth motion without requiring the intermediate intensity frame generation step. The method provides a new way to synthesize vision event datasets, overcome the scarcity of event datasets, and improve upon the currently quality-wise unusable text-to-video-to-events pipeline. Our method is able to synthesize gesture event streams from text prompts, outperforming previous text-to-events approaches that output synthetic events more similar to a static image. 
 We introduce a novel architecture and training technique for event frame autoencoders that can output sparse event frame outputs. In conjunction with a text-conditional diffusion U-net, we successfully developed a text-to-events pipeline. The classification accuracy of the generated sequences using a classifier that is trained on real DVS128 dataset ranges between 42\% to 92\%, depending on the gesture group. A classifier trained on both real and separately generated samples will likely show higher accuracy scores. The generative performance of the pipeline itself may be further improved with additional regularization terms when generating event frames. Future directions include training with more diverse annotated datasets so as to allow general text prompt capabilities that match real event statistics and realistic noise profiles targeting specific event sensors.


{
    \bibliographystyle{IEEEtran}
    \bibliography{main}
}

\end{document}